\documentclass[runningheads,a4paper]{llncs}
\usepackage{booktabs,caption,fixltx2e}
\usepackage[flushleft]{threeparttable}
\usepackage{graphicx}
\usepackage[usenames,dvipsnames]{pstricks}
\usepackage{makeidx} 
\usepackage{amsmath}
\usepackage{multirow}

\usepackage{amsfonts}%
\usepackage{amssymb}%
\usepackage[english,oldcommands,plain,lined]{algorithm2e}
\usepackage{graphics}
\usepackage{subfig}
\usepackage[labelfont=bf]{caption}
\usepackage[font={small}]{caption}
\usepackage{transparent}
\usepackage{wrapfig}
\usepackage{color}
\usepackage{breqn}
\usepackage{epstopdf}
\usepackage[flushleft]{threeparttable}

\usepackage{color, colortbl}
\definecolor{LightCyan}{rgb}{0.88,1,1}
 \usepackage{cite}
\usepackage[font=small,labelfont=bf]{caption}
\usepackage{enumitem}
\usepackage{subfig}
\usepackage{makecell}
\usepackage{url}
\usepackage{hyperref}
\usepackage{relsize}
\usepackage{bm,array}
\usepackage{epsfig}
\usepackage{pst-grad} 
\usepackage{pst-plot} 
\usepackage[space]{grffile} 
\usepackage{etoolbox} 
\usepackage{pgfplots}
\usepackage{capt-of}
\usepackage{nicefrac}
\makeatletter 
\patchcmd\Gread@eps{\@inputcheck#1 }{\@inputcheck"#1"\relax}{}{}
\makeatother

\pgfplotsset{every tick label/.append style={font=\Large}}

\begin{document}
\newcolumntype{C}{>{\centering\arraybackslash}p{2em}}
\newcolumntype{x}[1]{>{\centering\hspace{0pt}}m{#1}}
\newcommand{\tn}{\tabularnewline}
\setlength{\abovedisplayskip}{2.9pt}
\setlength{\belowdisplayskip}{2.9pt}
\mainmatter  

\title{Learning Robust Hash Codes for Multiple Instance Image Retrieval}

\titlerunning{Learning Robust Hash Codes for Multiple Instance Image Retrieval}

%
%
\author{Sailesh Conjeti\inst{1}, Magdalini Paschali\inst{1}, Amin Katouzian\inst{2} \and Nassir Navab\inst{1,3}}
\authorrunning{Under review in MICCAI 2017} 
\institute{
Computer Aided Medical Procedures, Technische Universit\"{a}t M\"{u}nchen, Germany.\\
\and
IBM Almaden Research Center, Almaden, USA.\\
\and
Computer Aided Medical Procedures, Johns Hopkins University, USA.
}



%
%

\toctitle{Learning Robust Hash Codes for Multiple Instance Image Retrieval}
\tocauthor{Authors' Instructions}
\maketitle

\begin{abstract}
In this paper, for the first time, we introduce a \textit{multiple instance} (MI) deep hashing technique for learning discriminative hash codes with weak bag-level supervision suited for large-scale retrieval. We learn such hash codes by aggregating deeply learnt hierarchical representations across bag members through a dedicated MI pool layer. For better trainability and retrieval quality, we propose a two-pronged approach that includes robust optimization and training with an auxiliary single instance hashing arm which is down-regulated gradually. We pose retrieval for tumor assessment as an MI problem because tumors often coexist with benign masses and could exhibit complementary signatures when scanned from different anatomical views. Experimental validations on benchmark mammography and histology datasets demonstrate improved retrieval performance over the state-of-the-art methods.
\end{abstract}

\section{Introduction}

In breast examinations, such as mammography, detected actionable tumors are further examined through invasive histology. 
Objective interpretation of these modalities is fraught with high inter-observer variability and limited reproducibility~\cite{nature}. In this context, a reference based assessment, such as presenting prior cases with similar disease manifestations (termed Content Based Image Retrieval (CBIR)) could be used to circumvent discrepancies in cancer grading. With growing sizes of clinical databases, such a CBIR system ought to be both scalable and accurate. Towards this, hashing approaches for CBIR are being actively investigated for representing images as compact binary codes that can be used for fast and accurate retrieval~\cite{iupui2,KSH,deepresidual}.  

\begin{figure}[t]
\includegraphics[width=\textwidth]{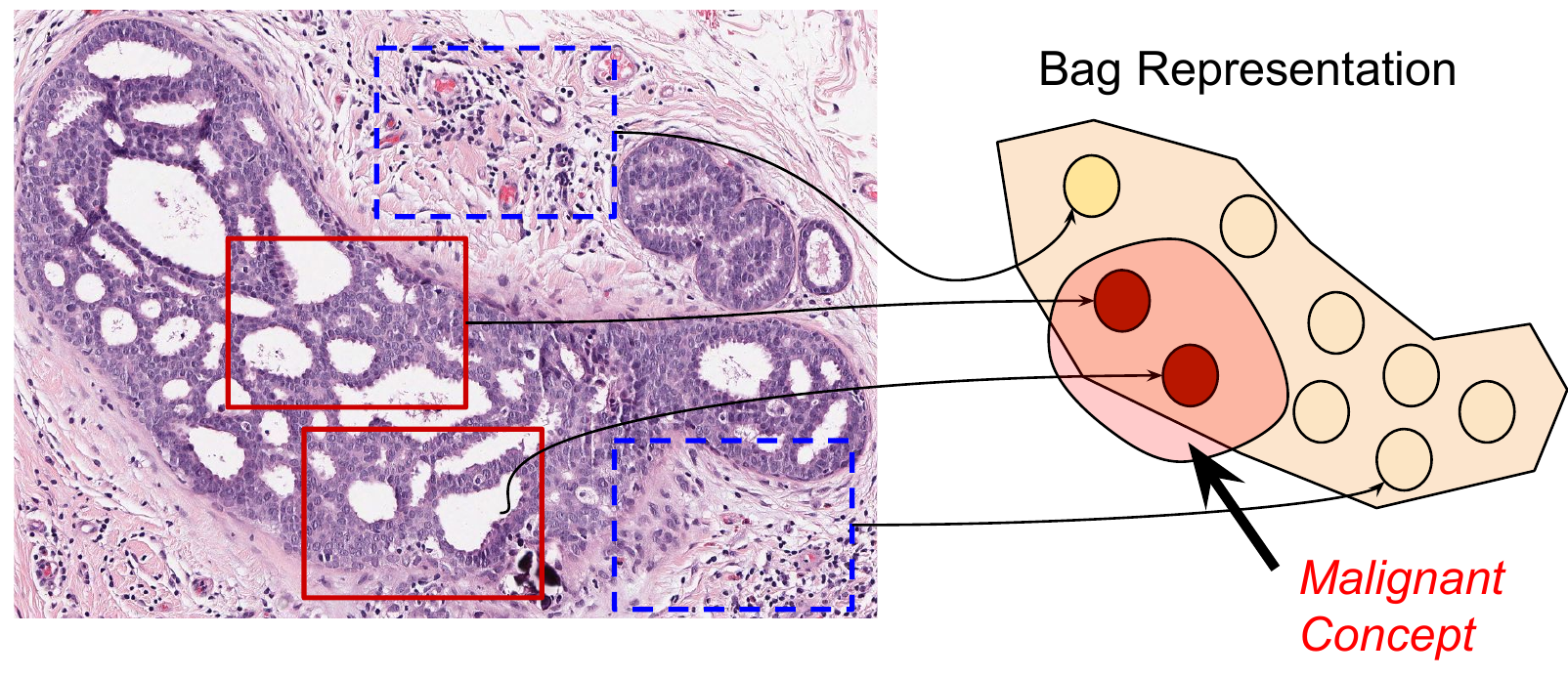}
\caption{Schematic representation of bag representation. Here, an example of a large region of interest ($\sim$ 6K $\times$ 4K) labeled as \textit{malignant} is shown wherein a few patches overlapping with the actual tumor represent the underlying malignant concept while proximal patches are potentially benign or less discriminative connective / lipidic tissues. It must be noted that these are not individually identified and only a bag-level weak annotation is available for learning.}
\label{fig:exampleBag}
\end{figure}

Malignant carcinomas are often co-located with
benign looking manifestations and suspect normal tissues. In such cases, describing the whole image with a single label is often inadequate for objective machine learning and alternatively requires expert annotations delineating the exact location of the 
region of interest. This argument extends to screening modalities like mammograms, where multiple anatomical views are acquired. In such scenarios, the status of the tumor is best represented to a CBIR system by constituting a bag of all associated images, thus veritably becoming multiple instance (MI) in nature. This is illustrated in Fig.~\ref{fig:exampleBag}. With this as our premise we present, for the first time, a novel deep learning based MI hashing method, termed as Robust Multiple Instance Hashing (RMIH). 

 Seminal works on shallow learning-based hashing include Iterative Quantization (ITQ)~\cite{ITQ}, Kernel Sensitive Hashing (KSH)~\cite{KSH} \textit{etc.} that propose a two-stage framework involving extraction of hand-crafted features followed by binarization. Yang \textit{et al.} extend these methods to MI learning scenarios with two variants: Instance Level MI Hashing (IMIH) and Bag Level MI Hashing (BMIH)~\cite{bmih}. However, these approaches are not end-to-end and are susceptible to semantic gap between features and associated concepts. Alternatively, deep hashing methods such as simultaneous feature learning and hashing (SFLH)~\cite{lai2015}, deep hashing networks (DHN)~\cite{zhuloss} and deep residual hashing (DRH)~\cite{deepresidual} to name a few, propose the learning of representations and hash codes in an end-to-end fashion, in effect bridging this semantic gap. It must be noted that all the above deep hashing works targeted single instance (SI) hashing scenarios and an extension to MI hashing was not investigated.
 
\begin{figure}[t]
\includegraphics[width=\textwidth]{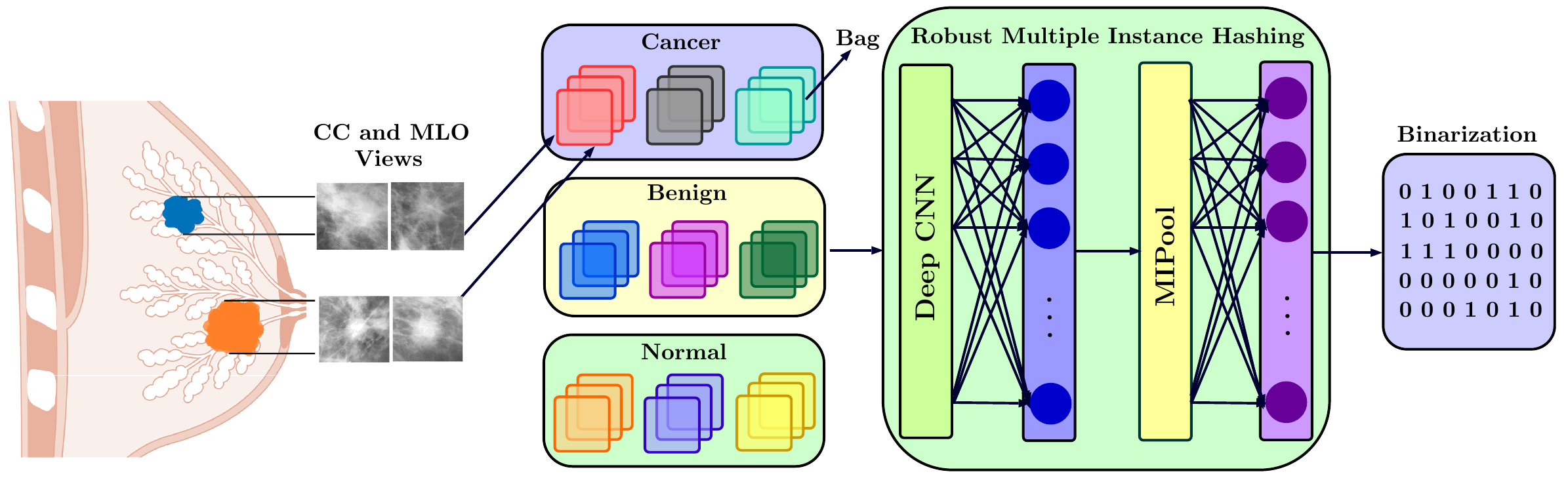}
\caption{Overview of RMIH for end-to-end generation of bag-level hash codes. Breast anatomy image is attributed to Cancer Research UK/Wikimedia Commons.}
\label{fig:overview}
\end{figure}


Earlier works on MI deep learning in computer vision include work by Wu \textit{et al.}~\cite{wu2015deep}, where the concept of an MI pooling (MIPool) layer is introduced to aggregate representations for multi-label classification. Yan \textit{et al.} leveraged MI deep learning for efficient body part recognition~\cite{bodypart}. Unlike MI classification that potentially substitutes the decision of the clinician, retrieval aims at presenting them with richer contextual information similar to the case at hand to facilitate decision-making.  
RMIH effectively bridges the two concepts for CBIR systems by combining the representation learning strength of deep MI learning with the potential for scalability arising from hashing. Within CBIR for breast cancer, notable prior art includes work on mammogram image retrieval by Jiang \textit{et al.}~\cite{mammogram1} and large-scale histology retrieval by Zhang \textit{et al.}~\cite{iupui2}. Both these works pose CBIR as an SI retrieval problem.
Contrasting with~\cite{mammogram1} and~\cite{iupui2}, within RMIH we create a bag of images
to represent a particular pathological case and generate a bag-level hash code, as shown in Fig.~\ref{fig:overview}. Our contributions in this paper include: \textbf{1)} introduction of a robust supervised retrieval loss for learning in presence of weak labels and potential outliers; \textbf{2)} propose training with an auxiliary SI arm with gradual loss trade-off for improved trainability; and \textbf{3)} incorporation of the MIPool layer to aggregate representations across variable number of instances within a bag, generating bag-level discriminative hash codes. 





\section{Methodology}
\begin{wrapfigure}[20]{r}{0.39\textwidth}
\vspace{-17pt}
\centering
\includegraphics[width=0.38\textwidth]{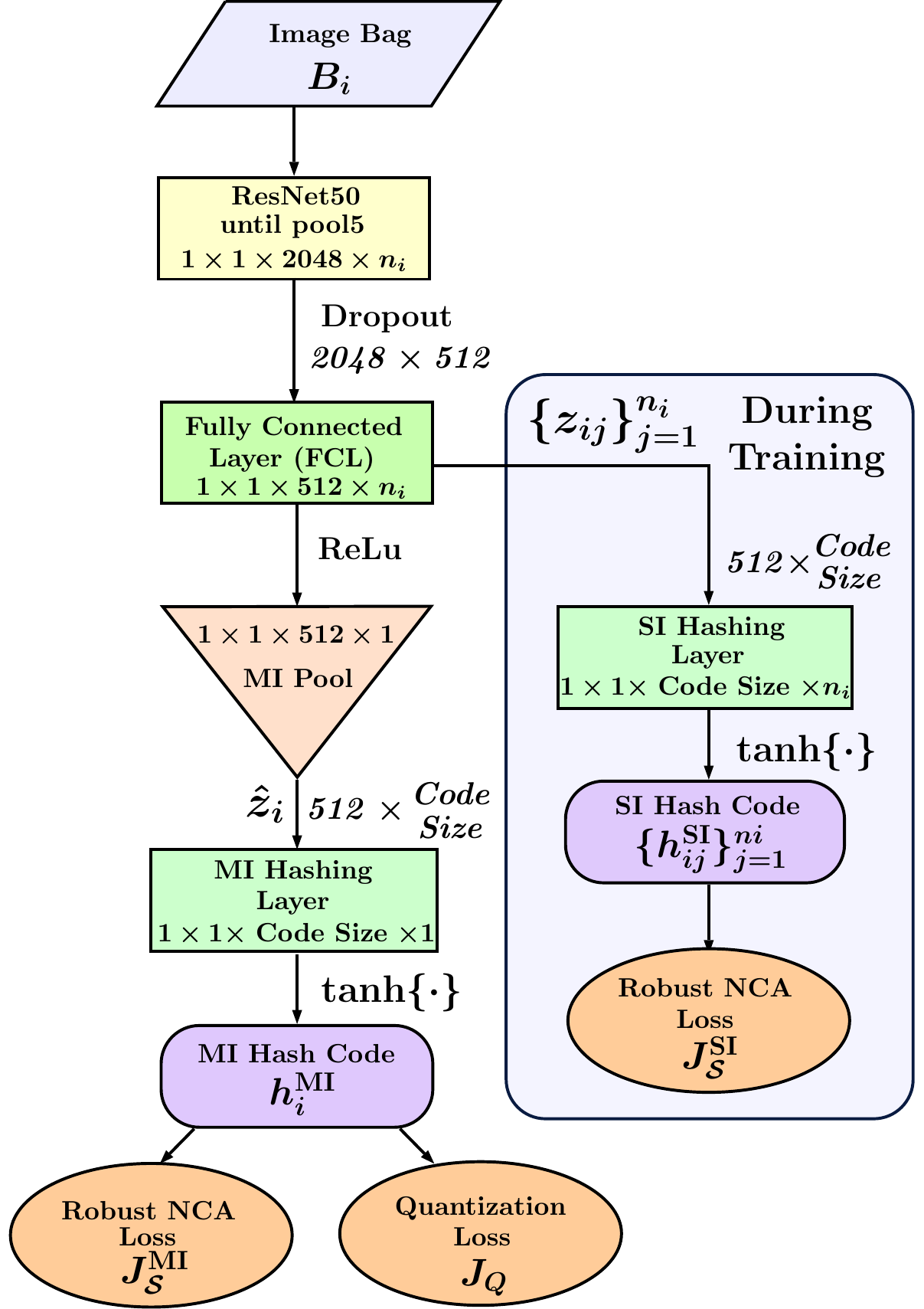}
\caption{RMIH Architecture with ResNet-50~\cite{residualHe} as the \textit{Deep CNN} model.}
\label{fig:architecture}
\end{wrapfigure}
Lets consider database $\mathcal{B} = \{B_1, \dots, B_{N_{B}}\}$ with $N_{B}$ bags. Each bag, $B_{i}$, with varying number ($n_{i}$) of instances ($I_{i}$) is denoted as $B_i = \{I_1,\dots,I_{n_{i}}\}$. We aim at learning $\mathcal{H}$ that maps each bag to a $K$-d Hamming space $\mathcal{H}:\mathcal{B} \rightarrow\{-1,1\}^K$, such that \textit{bags with similar instances and labels are mapped to similar codes}. For supervised learning of $\mathcal{H}$, we define a bag-level pairwise similarity matrix $\mathcal{S}^{\text{MI}}=\{s_{ij}\}^{N_{B}}_{ij=1}$, such that $s_{ij}=1$ if the bags are similar and zero otherwise. 
In applications, such as this one, where retrieval ground truth is unavailable
we can use classification labels as a surrogate for generating $\mathcal{S}^{\text{MI}}$.

%

\noindent
\textbf{Architecture}: As shown in Fig.~\ref{fig:architecture}, the proposed RMIH framework consists of a deep CNN terminating in a fully connected layer (FCL). Its outputs $\left \{ z_{ij} \right \}_{j=1}^{n_{i}}$ are fed into the MIPool layer to generate the aggregated representation $\hat{z}_{i}$ that is pooled ($\text{max}_{\forall j}\left \{ z_{ij} \right \}_{j=1}^{n_{i}}$, mean$(\cdot)$, \textit{etc.}
) across instances within the bag. 
$\hat{z}_{i}$ is an embedding in the space of the bags and is the input of a fully connected MI hashing layer. The output of this layer is squashed to $[-1, 1]$ by passing it through a tanh$\{\cdot\}$ function to generate $h_i^{\text{MI}}$, which is quantized to produce bag-level hash codes as $\mathbf{b}_{i}^{\text{MI}}=\ \text{sgn}\ (\mathbf{h}_{i}^{\text{MI}})$. The deep CNN mentioned earlier could
be a pretrained network, such as VGGF \cite{vgg}, GoogleNet~\cite{goingdeeper}, ResNet50 (R50)~\cite{residualHe} or an application specific network. 

During training of RMIH, we introduce an auxiliary SI hashing (aux-SI) arm, as shown in Fig.~\ref{fig:architecture}. It taps off at the FCL layer and feeds directly into a fully connected SI hashing layer with tanh$\{\cdot\}$ activation to generate instance level non-quantized hash codes, denoted as $\{h_{ij}^{\text{SI}}\}_{j=1}^{n_{i}}$. While training RMIH using backpropagation, the MIPool layer significantly sparsifies the gradients (analogous to using very high dropout while training CNNs), thus limiting the trainability of the preceding layers. The SI hashing arm helps to potentially mitigate this by producing auxiliary instance level gradients. 

\noindent
\textbf{Model Learning and Robust Optimization}: 
To learn similarity preserving hash codes, we propose a robust version of supervised retrieval loss based on neighborhood component analysis employed by~\cite{torralba}. The motivation to introduce robustness within the loss function is two-fold: (1) robustness induces immunity to potentially noisy
labels due to high inter-observer variability and limited reproducibility for the applications at hand~\cite{nature}; (2) it can effectively counter ambiguous label assignment while training with the aux-SI hashing arm. 

Given $\mathcal{S}^{\text{MI}}$, the robust supervised retrieval loss $J_{\mathcal{S}}^{\text{MI}}$ is defined as: 

\begin{equation}
{J_{\mathcal{S}}^{\text{MI}} = 1 - \frac{1}{N_{B}^{2}} \sum_{i,j = 1}^{N_{B}} s_{ij}p_{ij}}
\end{equation}
\noindent
where  $p_{ij}$ is the probability that two bags (indexed as $i$ and $j$) are neighbors. Given hash codes $\mathbf{h_i} = \left \{ h_{i}^{k} \right \}_{k=1}^{K}$ and $\mathbf{h_j}$, we define a bit-wise residual operation $r_{ij}$ as $r_{ij}^k = (h_i^k - h_j^k)$. We estimate $p_{ij}$ as:
\begin{equation}
p_{ij} = \frac{e^{-\mathcal{L}_{\text{Huber}}(\mathbf{h_i},\mathbf{h_j})}}{\sum_{i\neq l}^{N_{B}}e^{-\mathcal{L}_{\text{Huber}}(\mathbf{h_i},\mathbf{h_l})}},\text{  where } \mathcal{L}_{\text{Huber}}(\mathbf{h_i},\mathbf{h_j}) = \sum_{\forall k}\rho_k(r_{ij}^{k})
\end{equation}
\noindent
The Huber norm's robustness operation $\rho_k$ is defined as: 
\begin{equation}
\rho_k(r_{ij}^k) = \begin{cases}
\dfrac{1}{2}(r_{ij}^k)^2, &\text{if\ $\mid r_{ij}^k \mid \leqslant c_k$}\\
c_k \mid r_{ij}^k \mid -  \dfrac{1}{2} c_k^2, &\text{if $\mid r_{ij}^k \mid > c_k$}
\end{cases}
\label{eq:huber}
\end{equation}
\noindent
In Eq.~\eqref{eq:huber}, the tuning factor $c_k$ is estimated inherently from the data and is set to $c_k=1.345 \times \sigma_k$. The factor of $1.345$ is chosen to provide approximately 95\% asymptotic efficiency and $\sigma_k$ is a robust measure of bit-wise variance of $r_{ij}^k$. Specifically, $\sigma_k $ is estimated as $1.485$ times the median absolute deviation of $r_{ij}^k$ as empirically suggested in~\cite{robusthuber}. This robust formulation provides immunity to outliers during training by clipping their gradients. For training with the aux-SI hashing arm, we employ a similar robust retrieval loss $J_{\mathcal{S}}^{\text{SI}}$ defined over single instances with bag-labels assigned to member instances.

To minimize loss of retrieval quality due to quantization, we use a differentiable quantization loss $J_{Q} = \sum_{i=1}^M(\text{log}\ \text{cosh}(\mid\mathbf{h_i}\mid - \ \mathbf{1}))$ proposed in~\cite{zhuloss}. This loss also counters the effect of using continuous relaxation in definition of $p_{ij}$ over using Hamming distance. 
As a standard practice in deep learning, we also add an additional weight decay regularization term $R_{W}$, which is the Frobenius norm of the weights and biases, to regularize the cost function and avoid over-fitting.


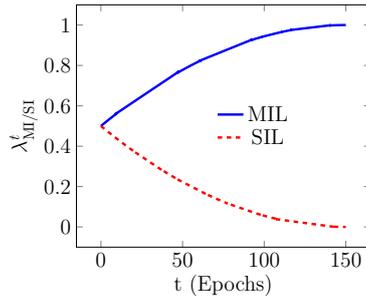
\begin{wrapfigure}[15]{r}[0pt]{0pt}
\resizebox{0.4\textwidth}{!}{
\begin{tikzpicture}
\begin{axis}[
	xlabel= \Large{t (Epochs)},ylabel= {\Large{$\lambda_{\text{MI/SI}}^t$}},ylabel shift = -1 pt,legend style={at={(0.46,0.5)},anchor=west,draw=none}]
\addplot[smooth,blue,ultra thick] coordinates {
(0,	0.500000000000000)
(1.00671140939597,	0.506688887887933)
(2.01342281879195,	0.513332732759786)
(3.02013422818792,	0.519931534615558)
(4.02684563758389,	0.526485293455250)
(5.03355704697987,	0.532994009278861)
(6.04026845637584,	0.539457682086392)
(7.04697986577181,	0.545876311877843)
(8.05369127516779,	0.552249898653214)
(9.06040268456376,	0.558578442412504)
(10.0671140939597,	0.564861943155714)
(11.0738255033557,	0.571100400882843)
(45.3020134228188,	0.756407369037431)
(46.3087248322148,	0.761069321201748)
(47.3154362416107,	0.765686230349984)
(48.3221476510067,	0.770258096482140)
(49.3288590604027,	0.774784919598216)
(50.3355704697987,	0.779266699698212)
(51.3422818791946,	0.783703436782127)
(52.3489932885906,	0.788095130849962)
(53.3557046979866,	0.792441781901716)
(54.3624161073826,	0.796743389937390)
(55.3691275167785,	0.800999954956984)
(56.3758389261745,	0.805211476960497)
(57.3825503355705,	0.809377955947930)
(58.3892617449664,	0.813499391919283)
(59.3959731543624,	0.817575784874555)
(60.4026845637584,	0.821607134813747)
(61.4093959731544,	0.825593441736859)
(62.4161073825503,	0.829534705643890)
(90.6040268456376,	0.921602630512139)
(91.6107382550336,	0.924237646952840)
(92.6174496644295,	0.926827620377461)
(93.6241610738255,	0.929372550786001)
(94.6308724832215,	0.931872438178460)
(95.6375838926175,	0.934327282554840)
(96.6442953020134,	0.936737083915139)
(97.6510067114094,	0.939101842259358)
(98.6577181208054,	0.941421557587496)
(99.6644295302014,	0.943696229899554)
(109.731543624161,	0.963965587135715)
(110.738255033557,	0.965744786270889)
(111.744966442953,	0.967478942389982)
(112.751677852349,	0.969168055492996)
(113.758389261745,	0.970812125579929)
(114.765100671141,	0.972411152650782)
(115.771812080537,	0.973965136705554)
(116.778523489933,	0.975474077744246)
(117.785234899329,	0.976937975766857)
(138.926174496644,	0.997274897527138)
(139.932885906040,	0.997747849195982)
(140.939597315436,	0.998175757848746)
(141.946308724832,	0.998558623485429)
(142.953020134228,	0.998896446106031)
(143.959731543624,	0.999189225710554)
(144.966442953020,	0.999436962298996)
(145.973154362416,	0.999639655871357)
(146.979865771812,	0.999797306427638)
(147.986577181208,	0.999909913967839)
(148.993288590604,	0.999977478491960)
(150,	1)
};
\addlegendentry{\Large{MIL}}

\addplot[smooth,color=red,dashed,ultra thick]
	coordinates {
(0,	0.500000000000000)
(1.00671140939597,	0.493311112112067)
(2.01342281879195,	0.486667267240214)
(3.02013422818792,	0.480068465384442)
(4.02684563758389,	0.473514706544750)
(5.03355704697987,	0.467005990721139)
(6.04026845637584,	0.460542317913608)
(7.04697986577181,	0.454123688122157)
(8.05369127516779,	0.447750101346786)
(9.06040268456376,	0.441421557587496)
(10.0671140939597,	0.435138056844286)
(11.0738255033557,	0.428899599117157)
(12.0805369127517,	0.422706184406108)
(13.0872483221477,	0.416557812711139)
(14.0939597315436,	0.410454484032251)
(15.1006711409396,	0.404396198369443)
(16.1073825503356,	0.398382955722715)
(17.1140939597315,	0.392414756092068)
(18.1208053691275,	0.386491599477501)
(19.1275167785235,	0.380613485879014)
(20.1342281879195,	0.374780415296608)
(21.1409395973154,	0.368992387730282)
(36.2416107382550,	0.287577136165038)
(37.2483221476510,	0.282509796855998)
(38.2550335570470,	0.277487500563038)
(39.2617449664430,	0.272510247286158)
(40.2684563758389,	0.267578037025359)
(41.2751677852349,	0.262690869780641)
(42.2818791946309,	0.257848745552002)
(43.2885906040268,	0.253051664339444)
(44.2953020134228,	0.248299626142967)
(45.3020134228188,	0.243592630962569)
(46.3087248322148,	0.238930678798252)
(47.3154362416107,	0.234313769650016)
(48.3221476510067,	0.229741903517860)
(64.4295302013423,	0.162717895590289)
(65.4362416107383,	0.158911760731499)
(66.4429530201342,	0.155150668888789)
(67.4496644295302,	0.151434620062159)
(68.4563758389262,	0.147763614251610)
(69.4630872483222,	0.144137651457142)
(70.4697986577181,	0.140556731678753)
(71.4765100671141,	0.137020854916445)
(72.4832214765101,	0.133530021170218)
(73.4899328859060,	0.130084230440070)
(74.4966442953020,	0.126683482726003)
(75.5033557046980,	0.123327778028017)
(76.5100671140940,	0.120017116346111)
(77.5167785234899,	0.116751497680285)
(78.5234899328859,	0.113530922030539)
(96.6442953020134,	0.0632629160848610)
(97.6510067114094,	0.0608981577406423)
(98.6577181208054,	0.0585784424125040)
(99.6644295302014,	0.0563037701004459)
(100.671140939597,	0.0540741408044683)
(101.677852348993,	0.0518895545245710)
(102.684563758389,	0.0497500112607541)
(103.691275167785,	0.0476555110130175)
(104.697986577181,	0.0456060537813612)
(105.704697986577,	0.0436016395657852)
(106.711409395973,	0.0416422683662898)
(107.718120805369,	0.0397279401828747)
(108.724832214765,	0.0378586550155399)
(109.731543624161,	0.0360344128642853)
(141.946308724832,	0.00144137651457144)
(142.953020134228,	0.00110355389396877)
(143.959731543624,	0.000810774289446425)
(144.966442953020,	0.000563037701004410)
(145.973154362416,	0.000360344128642831)
(146.979865771812,	0.000202693572361579)
(150,	0)

	};
\addlegendentry{\Large{SIL}}
\end{axis}
\end{tikzpicture}}
\vspace{-36pt}
\caption{\scriptsize{Weight Trade-off.}}\label{wrap-weights}
\end{wrapfigure}

\noindent
The following composite loss is used to train RMIH: 
\begin{equation}
J = \lambda_{\text{MI}}^{t} J_{\mathcal{S}}^{\text{MI}}+ \lambda_{\text{SI}}^{t} J_{\mathcal{S}}^{\text{SI}}+ \lambda_q J_{Q} + \lambda_w R_{W}
\end{equation}
where $ \lambda_{\text{MI}}^{t}$, $ \lambda_{\text{SI}}^{t}$, $\lambda_q$ and $\lambda_w$ are hyper-parameters that control the contribution of each of the loss terms. Specifically, $ \lambda_{\text{MI}}^{t}$ and $ \lambda_{\text{SI}}^{t}$ control the trade-off between the MI and SI hashing losses. The SI arm plays a significant role only in the early stages of training and can be traded off eventually to avoid sub-optimal MI hashing. For this we introduce a weight trade-off formulation that gradually down-regulates  $ \lambda_{\text{SI}}^{t}$, while simultaneously up-regulating $ \lambda_{\text{MI}}^{t}$. Here, we use $\lambda_{\text{SI}}^{t} = 1 - 0.5\left ( 1 - \nicefrac{t}{t_{\text{max}}} \right )^{2}$ and $\lambda_{\text{MI}}^{t} = 1 - \lambda_{\text{SI}}^{t}$, where $t$ is the current epoch and $t_{\text{max}}$ is the maximum number of epochs (see Fig.~\ref{wrap-weights}). We train RMIH with mini-batch stochastic gradient descent (SGD) with momentum. Due to potential outliers that can occur at the beginning of training, we scale $c_{k}$ up by a factor of 7 for $t = 1$ to allow a stable state to be reached. 
Specifically, the gradient of $J_{\mathcal{S}}^{(\cdot)}$ w.r.t. to $\mathbf{h}_{i}$ is derived as: 
\begin{align}
\frac{\partial J_{\mathcal{S}}^{(\cdot)}}{\partial \mathbf{h}_{i}} = &\left ( \sum_{l:s_{li} > 0} p_{li}\mathbf{\mathcal{L'_H}}(\mathbf{h}_l,\mathbf{h}_i) - \sum_{l \neq i}\left ( \sum_{q:s_{lq} > 0} p_{lq} \right ) p_{li}\mathbf{\mathcal{L'_H}}(\mathbf{h}_l,\mathbf{h}_i) \right ) \nonumber \\ 
- &\left ( \sum_{j:s_{ij} > 0} p_{ij}\mathbf{\mathcal{L'_H}}(\mathbf{h}_i,\mathbf{h}_j) - \sum_{j:s_{ij} > 0} p_{ij}\left ( \sum_{z \neq i} p_{iz}\mathbf{\mathcal{L'_H}}(\mathbf{h}_i,\mathbf{h}_z) \right ) \right ) 
\label{eq:derJSbig}
\end{align}
\noindent
where ${\mathcal{L_H'}}(\mathbf{h_i},\mathbf{h_j}) = \{ {\rho_k'}(r_{ij}^k)\}_{k=1}^{k}$. The derivative of the huber term ${\rho_k}'(r_{ij}^k)$ can be computed as:
\begin{equation}
{\rho_k'}(r_{ij}^k) = \begin{cases}
r_{ij}^k, &\text{if\ $\mid r_{ij}^k \mid \leqslant c_k$}\\
c_k\ \text{sgn}(r_{ij}^k), &\text{if $\mid r_{ij}^k \mid > c_k$}
\end{cases}
\end{equation}

Regarding the quantization loss function, the derivative can be computed by $\frac{\partial J_{Q}}{\partial \mathbf{h}_{i}} = \text{tanh}\left ( \left | \mathbf{h}_{i} \right | - \mathbf{1} \right )\text{sgn}\left ( \mathbf{h}_{i} \right )$. Having computed these gradients, we use backpropagation to compute the derivatives of the preceding layers.

\begin{figure}[t]
\centering
\begin{minipage}{\textwidth}
  \begin{minipage}[t]{0.6\textwidth}
\centering
\includegraphics[width=\textwidth]{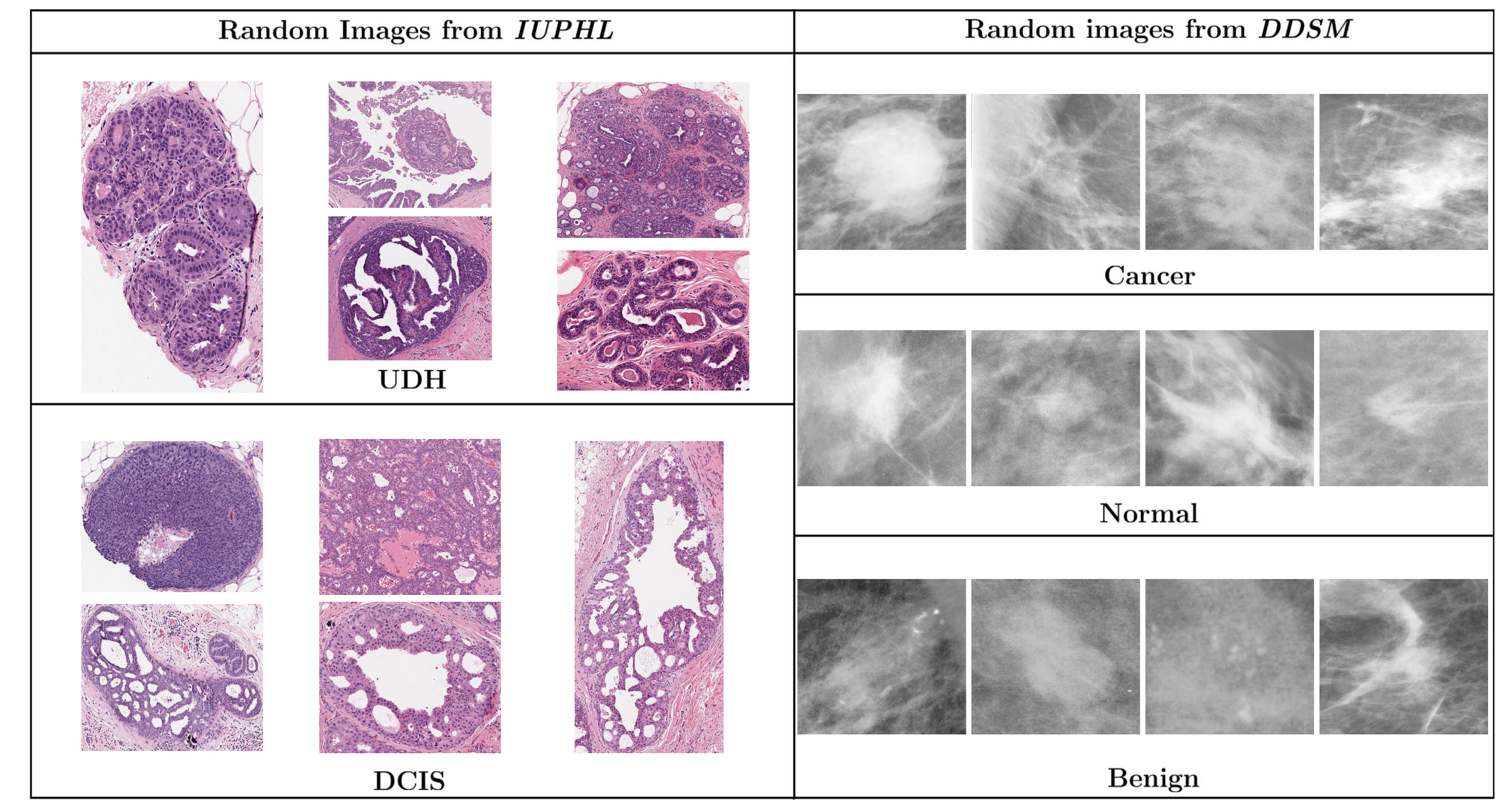}
  \end{minipage}
  \hfill
  \begin{minipage}[t]{0.36\textwidth}
\vspace{-3.4cm}
\captionof{figure}{Select images from the \textit{IPUHL} and \textit{DDSM} datasets to showcase the degree of anatomical variability within and across the classes.}
\label{fig:randomimages}
    \end{minipage}
  \end{minipage}
  \vspace{-0.7cm}
\end{figure}

\section{Experiments}

\textbf{Databases:} Clinical applicability of RMIH has been validated on two large scale datasets, namely, Digital Database for Screening Mammography (DDSM)~\cite{mammogram1,mammogram2} and a retrospectively acquired histology dataset from the Indiana University Health Pathology Lab (\textit{IUPHL})~\cite{iupui1,iupui2}. The \textit{DDSM} dataset comprises of 11,617 expert selected regions of interest (ROI) curated from 1861 patients. Multiple ROIs associated with a single breast from anatomical views constitute a bag (size: 1-12; median: 2), which has been annotated as normal, benign or malignant by expert radiologists.
A bag labeled \textit{malignant} could potentially contain multiple suspect normal and benign masses, which have not been individually identified. The \textit{IUPHL} dataset is a collection of 653 ROIs from histology slides from 40 patients (20 with precancerous ductal hyperplasia (UDH) and rest with ductal carcinoma \textit{in situ} (DCIS)) with ROI level annotations done by expert histopathologists. Due to high variability in sizes of these ROIs (upto 9K $\times$ 8K pixels), we extract multiple patches (of size $1024\times1024$) and populate a ROI-level bag (size: 1-15; median: 8). 
As cellular and nuclei level characteristics are important to distinguishing DCIS from UDH, it is not recommended to rescale these images to standard input sizes used by CNNs (typically, 244 $\times$ 224 in~\cite{vgg, residualHe, goingdeeper}). Fig.~\ref{fig:randomimages} illustrates select images from the two datasets to showcase anatomical variability within and across the constituent classes.
From both the datasets, we use patient-level 
non-overlapping splits to constitute the training (80\%) and testing (20\%) sets. 

\textbf{Model Settings and Validations}: To validate proposed contributions, namely robustness within NCA loss and trade-off from the aux-SI arm, we perform ablative testing with combinations of their baseline variants by fine-tuning multiple network architectures. Additionally, we compare RMIH against four state-of-the art methods: ITQ~\cite{ITQ}, KSH~\cite{KSH}, SFLH~\cite{lai2015} and DHN~\cite{zhuloss}. For a fair comparison, we use R50 for both SFLH and DHN, since as discussed later it performs the best. Since SFLH and DHN were originally proposed for SI hashing, we introduce additional MI variants by hashing through the MIPool layer.
For ITQ and KSH, we further create two comparative settings: \textbf{1)} Using IMIH~\cite{bmih} that learns instance-level hash codes followed by bag-level distance computation and \textbf{2)} Utilizing BMIH~\cite{bmih} using bag-level kernalized representations followed by binarization.
For IMIH and SI variants of SFLH, DHN and RMIH, given two bags  $B_p$ and $B_q$ with SI hash codes, say  $\mathcal{H}(B_q) = \{h_{q1},\dots,h_{qM}\}$ and $\mathcal{H}(B_p) = \{h_{p1},\dots,h_{pN}\}$, the bag-level distance is computed as: 
\begin{equation}
d(B_p,B_q) = \frac{1}{M}\sum_{i=1}^{M}(\min_{\forall j }\ \text{Hamming}(h_{pi},h_{qj})).
\label{eq:bagDist}
\end{equation}

All images were resized to $224 \times 224$ and training data were augmented
to create equally balanced classes. $ \lambda_{\text{MI}}^{t}$ and $ \lambda_{\text{SI}}^{t}$ were set assuming $t_{\text{max}}$ as 150 epoch; $\lambda_q$ and $\lambda_w$ were set at 0.05 and 0.001 respectively. The momentum term within SGD was set to 0.9 and batch size to 128 for \textit{DDSM} and 32 for \textit{IUPHL}. For efficient learning, we use an exponentially decaying learning rate initialized at 0.01. The RMIH framework was implemented in MatConvNet~\cite{matconvnet}. We use standard retrieval quality metrics: nearest neighbor classification accuracy (nnCA) and precision-recall (PR) curves to perform the aforementioned comparisons. The results (nnCA)
from ablative testing and comparative methods are tabulated in Table~\ref{wrap-tab:ablative} and Table~\ref{table:CM} respectively. Within Table~\ref{table:CM}, methods were evaluated at two different code sizes (16 bits and 32 bits). We also present the PR curves of select bag-level methods (32 bits) in Fig. \ref{fig:PRCurves}.

\section{Results and Discussion} \label{results}

\begin{figure}[t]
\resizebox{\textwidth}{!}{\begin{tabular}{|c|c|C|C|c|c|c|c|c|c|} \hline
\multicolumn{2}{|c|}{\multirow{2}{*}{\textbf{Method}}} & \multicolumn{2}{c|}{\thead{\textbf{Variants}}} & \multicolumn{3}{c|}{\thead{\textit{\textbf{DDSM}}}} & \multicolumn{3}{c|}{\thead{\textit{\textbf{IUPHL}}}}\\ \cline{3-10}
\multicolumn{2}{|c|}{\multirow{2}{*}{}} & \thead{R} &\thead{T} &\thead{\ VGGF$\ $} &\thead{\ R50$\ $} &\thead{\ GN$\ $ }&\thead{\ VGGF$\ $} &\thead{\ R50$\ $} &\thead{\ GN$\ $} \\ \hline
\multirow{5}{*}{\makecell{{\small \textbf{Ablative}}\\ {\small \textbf{Testing}}}} & A & $\circ$ & $\circ$ & 68.65 & 72.76  & 71.70 & 83.85  & 85.42  &  82.29  \\ \cline{2-10}
& B & $\circ$ & $\bullet$ & 75.38 & 77.34  & 72.92  & 85.94  & 90.10  & 88.02  \\ \cline{2-10}
& C & $\bullet$ & $\circ$  & 70.65 & 76.63 & 70.02 & 83.33  & 85.94  &  86.46  \\ \cline{2-10}
& D & $\circ$ & \tiny{$\blacksquare$} & 66.65 & 69.67 & 68.26 & 83.33  & 88.54  &  84.90  \\ \cline{2-10}
& E & $\bullet$ & \tiny{$\blacksquare$} & 67.05 & 76.59 & 72.84 & 84.38  & 89.58  &  85.42  \\\hline
\multicolumn{2}{|c|}{\textbf{RMIH-mean}} & $\bullet$ & $\bullet$ & 78.67 & 82.31 & 76.83  &  87.50 & 89.58  & \textbf{89.06} \\ \hline
\multicolumn{2}{|c|}{\textbf{RMIH-max}}& $\bullet$ & $\bullet$ & \textbf{81.21} & \textbf{85.68}  & \textbf{78.67}  & \textbf{91.67} & \textbf{95.83} & 88.02 \\ \hline
\multicolumn{2}{|c|}{\textbf{RMIH}($\lambda_q=0$)} & $\bullet$ & $\bullet$ & 75.34 & 79.88 & 73.06 & 87.50 & 89.58  &  88.51 \\ \hline
\multicolumn{2}{|c|}{\textbf{RMIH NB}}  &  $\bullet$ & $\bullet$ & 83.25 & 88.02 & 79.06  & 94.79  & 96.35 & 92.71 \\ \hline
\multirow{4}{*}{\textbf{Legend}}  
& \multicolumn{3}{r|}{\textbf{R(Robustness)}} & \multicolumn{6}{l|}{ $\circ=L_2,\  \bullet=L_{\text{Huber}}$}\\ 
& \multicolumn{3}{r|}{\makecell[r]{\textbf{T(Trade-off)}}} & \multicolumn{6}{l|}{\makecell[l]{$\circ=\text{\small{Equal weights,}}\ \bullet=\text{\small{Decaying SIL weights,}}$ \\ \scriptsize{$\blacksquare$} \small{$=\text{No SIL branch}$}}}\\ 
& \multicolumn{3}{r|}{\textbf{Networks}} & \multicolumn{6}{l|}{R50: ResNet50,\ GN: GoogleNet} \\  \hline
\end{tabular}}
\caption{Performance of ablative testing at code size of 16 bits. We report the nearest neighbor classification accuracy (nnCA) estimated over unseen test data. Letters A-E are introduced for easier comparisons, discussed in Section \ref{results}.}
\label{wrap-tab:ablative}
 \vspace{-0.5cm}
\end{figure}

\textbf{Effect of aux-SI Loss}: To justify using the aux-SI loss, we introduce a variant of RMIH without it (E in Table~\ref{wrap-tab:ablative}), which
leads to a significant decline of 3\% to 14\% 
in contrast to RMIH. This could be potentially attributed to the prevention of the gradient sparsification caused by the MIPool layer. From Table~\ref{wrap-tab:ablative}, we observe a 3\%-10\% increase in performance, comparing cases with gradual decaying trade-off (B)
against baseline setting ($\lambda^{t}_{\text{MI}} = \lambda^{t}_{\text{SI}} = 0.5$, A,C). 

\textbf{Effect of Robustness}: For robust-NCA, we compared against the original NCA formulation proposed in~\cite{torralba} (A,B,D in Table~\ref{wrap-tab:ablative}). Robustness helps handle potentially noisy MI labels, inconsistencies within a bag (like non-informative patches) and the ambiguity in assigning SI labels. Comparing the effect of robustness  
for baselines sans the SI hashing arm (D \textit{vs.} E) 
we observe marginally positive improvement across the architectures and datasets, with a substantial 7\% in ResNet50 for \textit{DDSM}. Robustness contributes more with the addition of the aux-SI hash arm (proposed \textit{vs.} E)
with improved performance in the range of 4\%-5\% across all settings. This observation further validates our prior argument. 

\textbf{Effect of Quantization}: To assess the effect of quantization, we define two baselines: (1) setting $\lambda_{q} = 0$ and (2) using non-quantized hash codes for retrieval (RMIH - NB). The latter potentially acts as an upper bound for performance evaluation. From Table~\ref{wrap-tab:ablative}, we observe a consistent increase in performance by margins of 3\%-5\% if RMIH is learnt with an explicit quantization loss to limit the associated error. It must also be noted that comparing with RMIH - NB, there is only a marginal fall in performance (2\%-4\%), which is desired. 

Comparing max \textit{vs.} mean MI Pool variants, we observe that max achieves marginally better performance, since it is more selective than mean, which is particularly important in cases of detecting malignancy. 

\begin{figure}[t]
\includegraphics[width=\textwidth]{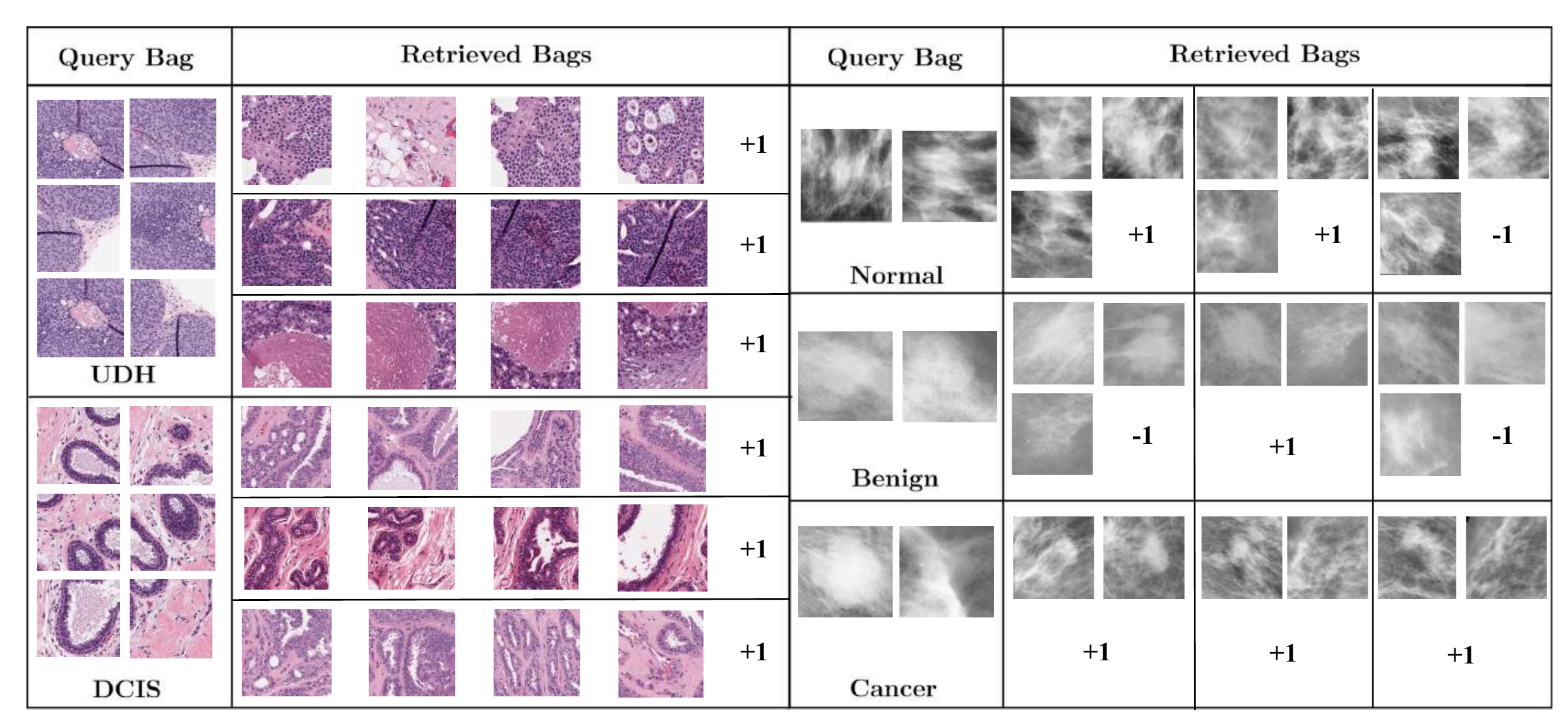}
\caption{Retrieval results for RMIH at code size 16 bits. +1 indicates retrieval from class consistent with query and -1 indicates otherwise.}
\label{fig:retrievedimages}
\vspace{-8pt}
\end{figure}

\begin{figure}[t]
\centering
\begin{minipage}{\textwidth}
  \begin{minipage}[t]{0.7\textwidth}
\centering
\includegraphics[width=\textwidth]{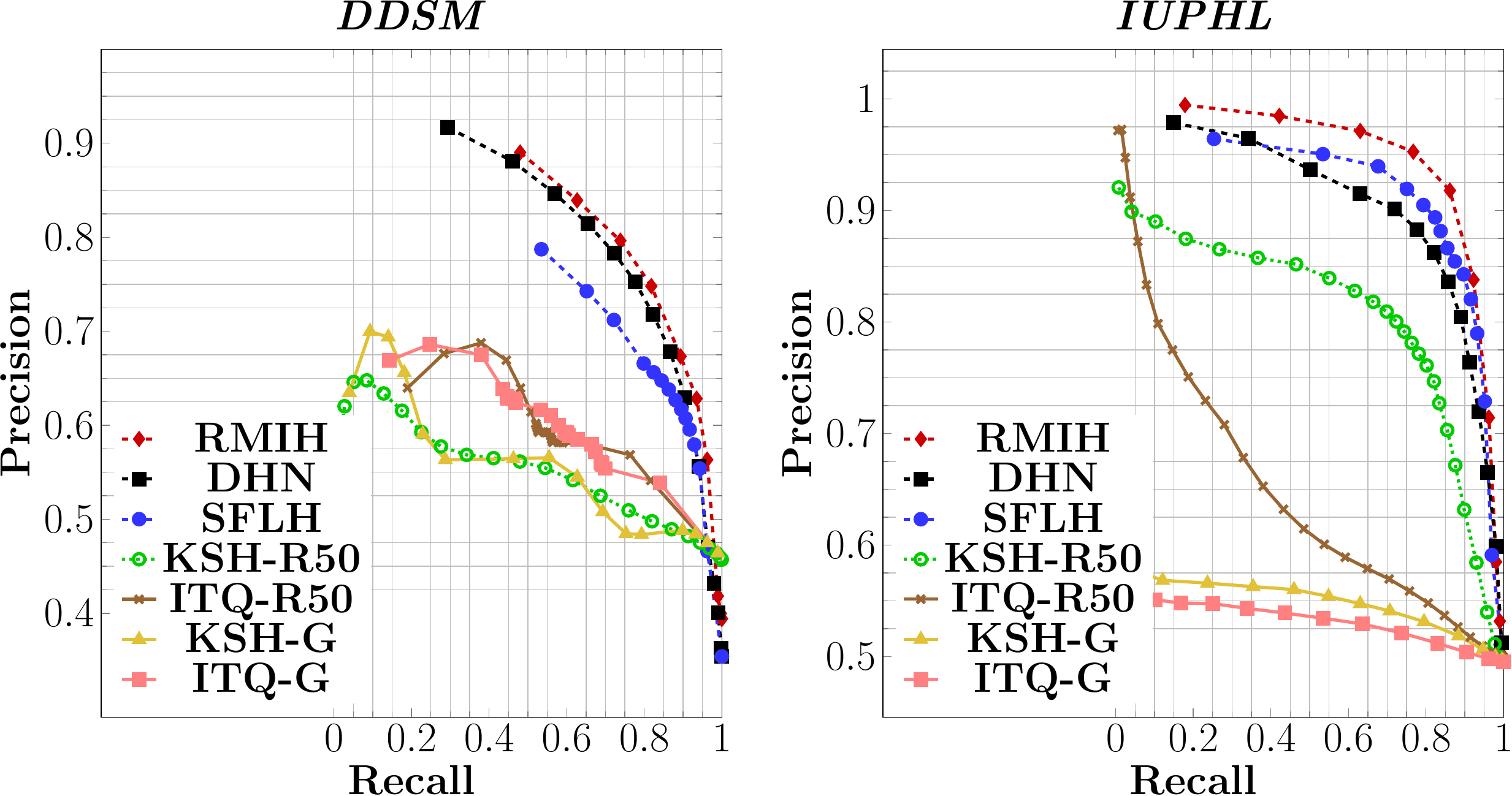}
  \end{minipage}
  \hfill
  \begin{minipage}[t]{0.26\textwidth}
\vspace{-2.7cm}
\captionof{figure}{PR curves for \textit{DDSM} and \textit{IUPHL} datasets at code size of 32.}
\label{fig:PRCurves}
    \end{minipage}
  \end{minipage}
  \vspace{-0.7cm}
\end{figure}

As a whole, the two-pronged proposed approach, including robustness and trade-off, along with quantization loss delivers the highest performance, proving that RMIH is able to learn effectively, despite the ambiguity induced by the SI hashing arm. Fig.~\ref{fig:retrievedimages} demonstrates the retrieval performance of RMIH on the target databases. For \textit{IUPHL}, the retrieved images are semantically similar to the query as consistent anatomical signatures
are evident in the retrieved neighbors. For \textit{DDSM}, in the cancer and normal cases the retrieved neighbors are consistent, however it is hard to distinguish between benign and malignant. The retrieval  time for a single query for RMIH was observed at 31.62 ms (for \textit{IUPHL}) and 17.48 ms (for \textit{DDSM}) showing potential for fast and scalable search.

\begin{wraptable}[18]{L}{6.0cm}
\vspace{-0.4cm}
    \centering
   \resizebox{5.9cm}{!}{\begin{tabular}{x{0.4cm}|x{2.2cm}|x{1cm}|x{0.5cm}|x{1cm}|x{1cm}|x{1.2cm}|x{1cm}|} 
\multicolumn{2}{c|}{\multirow{2}{*}{\textbf{Method}}} & \multirow{2}{*}{\thead{\textbf{A/F}}} & \multirow{2}{*}{\thead{\textbf{L}}} & \multicolumn{2}{c|}{\thead{\textit{\textbf{DDSM}}}} & \multicolumn{2}{c|}{\thead{\textit{\textbf{IUPHL}}}} \tn \cline{5-8}  
\multicolumn{2}{c|}{} & & & 16-bit &  32-bit &  16-bit &  32-bit \tn \cline{1-8}
\multirow{8}{*}{\rotatebox[origin=c]{90}{\textbf{Shallow}}} & \multirow{4}{*}{ITQ \cite{ITQ}} & R50 & $\circ$ & 66.35 & 67.71 & 78.58 &  80.28  \tn \cline{3-8}
& & R50 & $\bullet$ & 64.56 & 71.98 & 89.58 & 79.69 \tn \cline{3-8} 
& & G & $\circ$ & 65.22 & 66.55 & 51.79  & 51.42    \tn  \cline{3-8}
&  & G & $\bullet$ & 59.73 & 61.03 &  57.29 &  58.85  \tn \cline{2-8}
& \multirow{4}{*}{KSH \cite{KSH}} &  R50 & $\circ$  & 61.88 & 64.81 & 87.74 & 86.51 \tn \cline{3-8}
& &  R50 & $\bullet$ & 59.81 & 72.17 & 70.83 & 80.21 \tn \cline{3-8}
& & G & $\circ$  & 60.50 & 61.91  & 57.36  & 57.83 \tn \cline{3-8}
& & G & $\bullet$ & 55.34 & 55.67 & 60.94 &  58.85 \tn \hline
\multirow{6}{*}{\rotatebox[origin=c]{90}{\textbf{Deep}}} & \multirow{2}{*}{SFLH \cite{lai2015}} & R50  & $\circ$  & 73.54 & 77.46 & 83.33 & 85.94 \tn \cline{3-8}
&  &  R50M & \tiny{$\blacksquare$} & 71.98 & 75.93 & 85.42 & 88.54  \tn \cline{2-8}
& \multirow{2}{*}{DHN \cite{zhuloss}} &  R50 & $\circ$ & 65.64 & 74.79 & 82.29 & 86.46  \tn \cline{3-8}
&  &  R50M & \tiny{$\blacksquare$} & 72.88 & 80.43 & 88.02 & 90.62  \tn  \cline{2-8}
& RMIH-SIL &  R50 & $\circ$ & 76.02 & 78.37 & 87.92 & 88.58 \tn  \cline{2-8}
& \textbf{RMIH} & R50M  & \tiny{$\blacksquare$} & \textbf{85.68} & \textbf{89.47} & \textbf{95.83} & \textbf{93.23}  \tn \hline
\multirow{3}{*}{\rotatebox[origin=c]{90}{\textbf{Legend}}} & \makecell*[r]{\textbf{A/F:}} & \multicolumn{6}{l|}{\makecell*[l]{\small{\textbf{A}: Architecture,\ \textbf{F}: Features} \tn \scriptsize{R50: ResNet50, R50M: ResNet50$+$MIPool, G: GIST}}} \tn \cline{2-8}
& \makecell*[r]{\textbf{L:}} & \multicolumn{6}{l|}{\makecell*[l]{$\circ=\text{IMIH}$,\ $\bullet=\text{BMIH}$,\ \footnotesize{$\ \blacksquare$\ }=\small{\ End-to-end}}} \tn \hline
\end{tabular}}
\caption{Results of comparison with state-of-the art hashing methods. }\label{table:CM}
\end{wraptable}
\noindent
\textbf{Comparative Methods}
In the contrastive experiments against ITQ and KSH, hand-crafted GIST \cite{gist} features underperformed significantly, while the improvement with the R50 features ranged from 5\%-30\%. However, RMIH still performed 10\%-25\% better, proving that even if deep learnt features severely boost the performance, the shallow methods cannot fully breach the gap to the deep ones.
Comparing the SI with the MI variations of DHN, SFLH and RMIH, it is observed that the performance improved in the range of 3\%-11\%, suggesting that end-to-end learning of MI hash codes is preferred over two-stage hashing \textit{i.e.} hashing at SI level and comparing at bag level with Eq.~\eqref{eq:bagDist}. However, RMIH fares comparably better than both the SI and MI versions of SFLH and DHN, owing to the robustness of the proposed retrieval loss function to potentially noisy labels and inconsistent instances within bags (also evident from PR curves in Fig. \ref{fig:PRCurves}). In all fairness, the concepts of training with aux-SI hashing arm with gradual trade-off and robustness could be potentially adapted to SFLH and DHN to improve their MI hashing performance.  
As also seen from the associated PR curves in Fig. \ref{fig:PRCurves}, the performance gap between shallow and deep hashing methods remains significant despite using R50 features. Comparative results strongly support our premise that end-to-end learning of MI hash codes is preferred over conventional two-stage approaches.

\section{Conclusion}
In this paper, for the first time, we proposed an end-to-end deep robust hashing framework, termed RMIH, for retrieval under a multiple instance setting. We incorporate the MIPool layer to aggregate representations across instances to generate a bag-level discriminative hash code. We introduce the notion of robustness into our supervised retrieval loss and improve the trainability of RMIH by utilizing an aux-SI hashing arm regulated by a trade-off. Extensive validations and ablative testing on two public breast cancer datasets
demonstrate the superiority of RMIH and its potential for future extension to other MI applications. 
\vspace{-0.15cm}


\begin{thebibliography}{21}

\bibitem{nature} Duijm LEM, Louwman MWJ, Groenewoud JH, van de Poll-Franse LV, Fracheboud J, Coebergh JW. Inter-observer variability in mammography screening and effect of type and number of readers on screening outcome. In BJC 2009, pp. 901--907.

\bibitem{KSH} Liu W, Wang J, Ji R, Jiang YG, Chang SF. Supervised hashing with kernels. In CVPR 2012, pp. 2074--2081, IEEE.

\bibitem{deepresidual} Conjeti S, Guha Roy A, Katouzian A, Navab N. Deep Residual Hashing. arXiv:1612.05400, 2016.

\bibitem{iupui2} Zhang X, Liu W, Dundar M, Badve S, Zhang S. Towards large-scale histopathological image analysis: hashing-based image retrieval. In TMI 2015, IEEE.

\bibitem{ITQ} Gong Y, Lazebnik S. Iterative quantization: A procrustean approach to learning binary codes. In CVPR 2011, pp. 817--824, IEEE.

\bibitem{bmih} Yang Y, Xu X, Wang X, Guo S, Cui L. Hashing Multi-Instance Data from Bag and Instance Level. In APWeb: LNCS, vol. 9313, pp. 437--448, Springer 2015.

\bibitem{lai2015}
Lai H, Pan Y, Liu Y, Yan S. Simultaneous feature learning and hash coding with deep neural networks. In CVPR 2015, pp. 3270-3278.

\bibitem{zhuloss} Zhu H, Long M, Wang J, Cao Y. Deep Hashing Network for Efficient Similarity Retrieval. In AAAI 2016.

\bibitem{wu2015deep} Wu J, Yu Y, Huang C, Yu K. Multiple Instance Learning for Image Classification and Auto-Annotation. In CVPR 2015, pp. 3460--3469.

\bibitem{bodypart} Zhennan Y, Yiqiang Z, Zhigang P, Shu L, Shinagawa Y, Shaoting Z, Metaxas DN, Xiang SZ. Multi-Instance Deep Learning: Discover Discriminative Local Anatomies for Bodypart Recognition. In Trans Med Imaging 2016, pp. 1332--1343, IEEE.

\bibitem{mammogram1} Jiang M, Zhang S, Li H, Metaxas DN. Computer-aided diagnosis of mammographic masses using scalable image retrieval. In TBME 2015, vol. 62, pp. 783--792.

\bibitem{vgg} Chatfeld K, Simonyan K, Vedaldi A, Zisserman A. Return of the devil in the details: Delving deep into convolutional nets. arXiv:1405.3531, 2014.

\bibitem{goingdeeper} Szegedy C, Liu W, Jia Y, Sermanet P, Reed SE, Anguelov D, Erhan D, Vanhoucke V, Rabinovich A. Going Deeper with Convolutions In CVPR 2015, pp. 1--9.

\bibitem{residualHe} He K, Zhang X, Ren S, Sun J. Deep Residual Learning for Image Recognition. In CVPR 2016, pp. 770--778, IEEE Computer Society.

\bibitem{torralba} Torralba A, Fergus R, Weiss Y. Small codes and large image databases for recognition. In CVPR 2008, pp. 1--8, IEEE.

\bibitem{robusthuber} Huber PJ. Robust Statistics. In International Encyclopedia of Statistical Science 2011, pp. 1248--1251, Springer Berlin Heidelberg.

\bibitem{mammogram2} Heath M, Bowyer K, Kopans D, Kegelmeyer Jr WP, Moore R, Chang K, Munishkumaran S. Current status of the digital database for screening mammography. In Digital Mammography 1998, pp. 457--460, Springer Netherlands.

\bibitem{iupui1} Badve S, Bilgin G, Dundar M, Gürcan MN, Jain RK, Raykar VC, Sertel O. Computerized Classification of Intraductal Breast Lesions Using Histopathological Images. In Biomed. Engineering 2011, vol. 58, pp. 1977--1984, IEEE.

\bibitem{matconvnet} Vedaldi A, Lenc K. Matconvnet: Convolutional neural networks for matlab. In ACM Int. Conf. on Multimedia 2015, pp. 689-692, ACM.

\bibitem{gist} Oliva A, Torralba A. Modeling the Shape of the Scene: A Holistic Representation of the Spatial Envelope. In IJCV 2001, vol. 42, pp 145–-175.






















\end{thebibliography}
\end{document}